\documentclass{article}
\usepackage{spconf,amsmath,graphicx}

\ninept

\title{Multi-modal Automated Speech Scoring using Attention Fusion}
\name{Manraj Singh Grover$^1$, Yaman Kumar$^1$, Sumit Sarin$^2$, Payman Vafaee$^{3,4}$, Mika Hama$^4$, Rajiv Ratn Shah$^1$}
\address{$^1$Indraprastha Institute of Information Technology, New Delhi, India\\$^2$Netaji Subhas Institute of Technology, New Delhi, India\\$^3$Teachers College, Columbia University, New York, United States\\$^4$Second Language Testing Inc., Princeton, United States\\
\texttt{manrajg@iiitd.ac.in, yamank@iiitd.ac.in, rajivratn@iiitd.ac.in},\\ \texttt{sumits.ic.16@nsit.net.in, pv2203@tc.columbia.edu, mika.hama@2lti.com}}
\begin{document}
%
\maketitle
\begin{abstract}
In this study, we propose a novel multi-modal end-to-end neural approach for automated assessment of non-native English speakers' spontaneous speech using attention fusion. The pipeline employs Bi-directional Recurrent Convolutional Neural Networks and Bi-directional Long Short-Term Memory Neural Networks to encode acoustic and lexical cues from spectrograms and transcriptions, respectively. Attention fusion is performed on these learned predictive features to learn complex interactions between different modalities before final scoring. We compare our model with strong baselines and find combined attention to both lexical and acoustic cues significantly improves the overall performance of the system. Further, we present a qualitative and quantitative analysis of our model.
\end{abstract}
\begin{keywords}
automated speech scoring, spontaneous speech, end-to-end, multi-modal, attention fusion
\end{keywords}

\section{Introduction}
Automated Scoring (AS), in general, refers to the act of using computers to convert examinee's performance on standardized tests to some performance indicators. The questions types can range from multiple choice questions to written and spoken language assessment. AS systems are expected to pick up relevant signals from examinee's response and transform them to metrics which accurately infer their ability \cite{kumar2019get}. The task of speech scoring, specifically, is the task of standardized assessment of speaking proficiency for speakers of a language. Such assessments can then inform important decisions like the hiring of candidates for a company or admission to schools, judging academic proficiency levels, etc. The history of AS systems is quite old with its humble beginnings in 1964 when Page and colleagues \cite{page1966imminence, page1994computer} started scoring essay on punch cards.

The earlier systems judged the performance of L2 speakers by making them recite a written text \cite{witt2000phone}. These recitals were compared with the speech of native speakers, and scores were produced accordingly. They gradually became more complex and started including handcrafted features for measuring features like pronunciation, fluency, stress, intonation and content \cite{doi:10.1002/ets2.12198}. Although this worked in practice, yet this type of testing was not able to judge complex features like opinion formation,  argument depth, structure of the prose, \textit{etc}. With advances in machine learning and ASR, the type of testing changed from just recital to more complex forms involving open-ended questions \cite{taghipour2016neural}. Lately, studies have shown end-to-end systems performing better when compared to feature-based systems for this type of testing. In addition, such systems alleviate the development effort needed for scoring these responses \cite{taghipour2016neural,chen2018end,alikaniotis2016automatic,8683717}. While these black box systems allow automated extraction of meaningful representations, very few studies, particularly in speech scoring, share insights of the predictions made. In this paper, we present a novel multi-modal end-to-end neural pipeline for automated oral proficiency assessment of L2 English speakers. We leverage a deep-learning-based ASR system for transcription and Bi-directional Recurrent Convolutional Neural Networks (BDRCNN) and Bi-directional Long Short-Term Memory Neural Networks (BDLSTM) to learn lexical and acoustic feature representations. We perform attention fusion \cite{Hori_2017_ICCV} on these features to learn complex interactions among these features. For analysis, we leverage learned attention weights to understand the important parts of the response.



\begin{figure*}[t]
  \centering
  \includegraphics[width=0.9\linewidth]{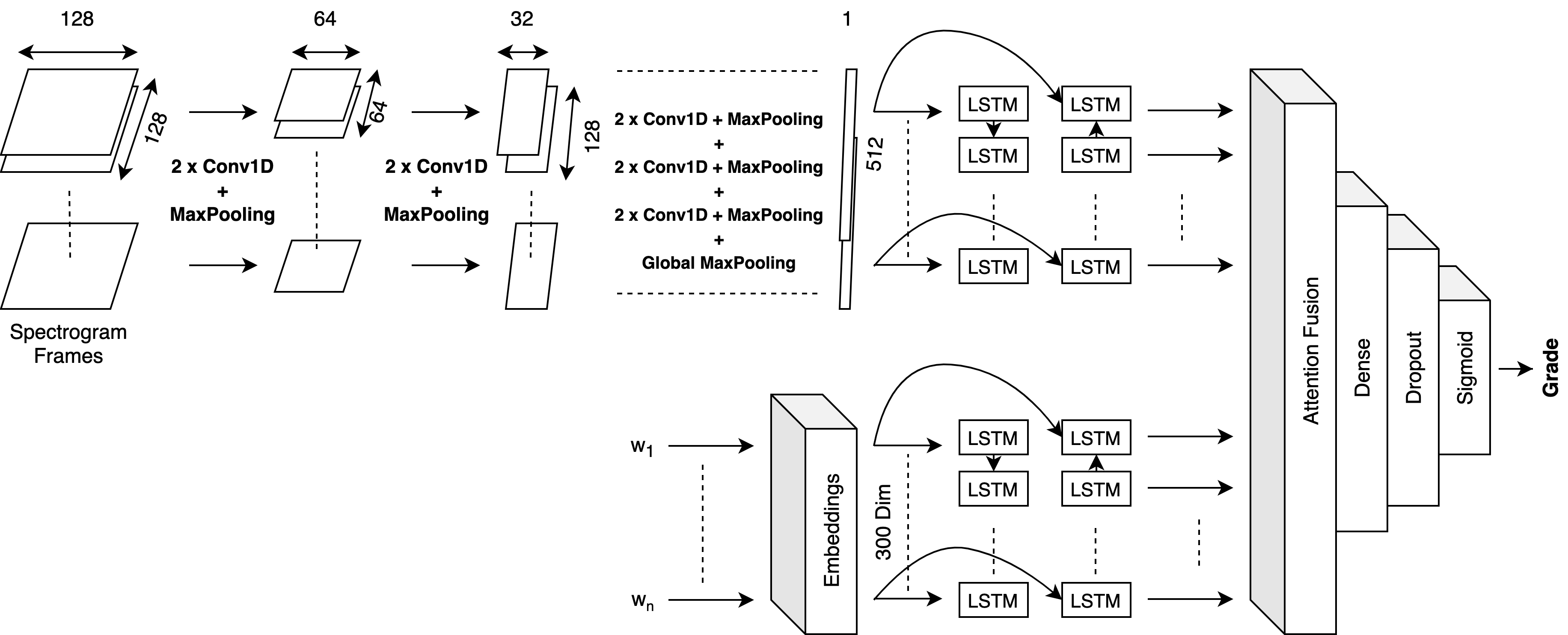}
  \caption{Multi-modal with Attention Fusion Speech Grading Architecture.}
  \label{fig:architecture}
\end{figure*}

\begin{table*}[thb]
  \caption{Statistics of the dataset. P: Prompt Number, \#R: Number of responses, D: Difficulty level, Sz: Average Response Size (Duration in seconds, Length in number of word tokens), and DS: Distribution of Scores (prefixes `L' means Low and `H' means High).}
  \label{tab:dataset}
  \centering
  \small
  \begin{tabular}{|*{10}{c|}}
    \hline 
    \textbf{P} &
    \textbf{\#R} &
    \textbf{D} &
    \multicolumn{2}{c|}{\textbf{Sz}} &
    \multicolumn{5}{c|}{\textbf{DS}} \\
    \cline{4-10}
    &&& \textbf{Duration} & \textbf{Length} & \textbf{A2} & \textbf{LB1} & \textbf{HB1} & \textbf{LB2} & \textbf{HB2} \\
    \hline
    1 & 8008 & B1 & 57.7 & 100 & 279 & 1575 & 6154 & - & -\\
    2 & 8160 & B1 & 58.7 & 108 & 546 & 3122 & 4492 & - & -\\
    3 & 8169 & B2 & 81.5 & 149 & 121 & 672 & 3535 & 3738 & 103\\
    4 & 8137 & C1 & 104.2 & 180 & 123 & 729 & 3573 & 3601 & 111\\
    5 & 8100 & C1 & 106.1 & 196 & 112 & 557 & 3061 & 4216 & 154\\
    6 & 8158 & B1 & 55.9 & 110 & 121 & 1045 & 6992 & - & - \\
    \hline
  \end{tabular}
\end{table*}

\section{Deep Learning based Scoring Models}
\label{scoringmodels}

\subsection{Bi-directional Recurrent CNN (for audio)}
\label{bdrcnnattn}

An important aspect of scoring rubric is the delivery of a spoken response. It helps in measuring speech quality, pronunciation, fluency, stress, and intonation \cite{doi:10.1002/ets2.12198}. To capture this acoustic information of audio samples, we propose a Recurrent Convolutional Neural Network (RCNN) architecture. CNN is able to extract local features, whilst the RNN is able to summarise the long temporal information \cite{7952585}. Such an architecture proves beneficial to us because of the nature of our data; which consists of long monologues (ranging from 60 to 120 seconds), and have long term interdependencies present in it.

We split the log-scaled mel spectrograms of the audio samples along the temporal dimension into frames of fixed size (discussed in Section~\ref{setup}). Each frame is passed to 5 sets of CNN consisting of 2 layers of 1D convolutions followed by max-pooling with the number of filters doubling after every set. Then, we apply global max pooling to each frame. The output vectors obtained from this network are then fed into the RNN component, which in our case is a BDLSTM allowing us to capture the sequential structure of audio, taking into account both past and future speech production. 

\subsection{Bi-directional Long Short-Term Memory (for text)}

It is also essential that the content of the spoken response is relevant to the topic and appropriate. Given the nature of our data, we use response-based content scoring, where we compare the similarity of response with previously scored responses. This approach is popular in both automated scoring of essays \cite{taghipour2016neural,AAAI1816431} as well as speech scoring \cite{chen2018end} as it allows capturing variations in responses comprehensively.

To achieve this, we generate transcripts of spoken responses using a non-native ASR system. These transcripts are pre-processed, and words are mapped to their word embeddings using an embedding layer. This embedding layer is initialized with 300-D glove embeddings \cite{pennington2014glove} trained on Wikipedia and is optimized during training. Similar to the acoustic model, we use BDLSTM network here to capture sequential the structure of words and learn the development of content at different score levels.

\begin{table*}[h]
  \caption{Quadratic Kappa Score and Mean Squared Error across prompts. ($^\dagger$ indicates threshold optimized models)}
  \label{tab:qwk}
  \small
  \centering
  \begin{tabular}{|c *{5}{|c c} | c c |}
    \hline
    \multicolumn{1}{|c}{\textbf{}} & 
    \multicolumn{2}{|c}{\textbf{Prompt 1}} & 
    \multicolumn{2}{|c}{\textbf{Prompt 2}} & 
    \multicolumn{2}{|c}{\textbf{Prompt 3}} & 
    \multicolumn{2}{|c}{\textbf{Prompt 4}} & 
    \multicolumn{2}{|c}{\textbf{Prompt 5}} & 
    \multicolumn{2}{|c|}{\textbf{Prompt 6}} \\
    \hline
    \multicolumn{1}{|c}{\textbf{Model}} & 
    \multicolumn{1}{|c}{\textbf{QWK}} & 
    \multicolumn{1}{c}{\textbf{MSE}} & 
    \multicolumn{1}{|c}{\textbf{QWK}} & 
    \multicolumn{1}{c}{\textbf{MSE}} &
    \multicolumn{1}{|c}{\textbf{QWK}} & 
    \multicolumn{1}{c}{\textbf{MSE}} &
    \multicolumn{1}{|c}{\textbf{QWK}} & 
    \multicolumn{1}{c}{\textbf{MSE}} &
    \multicolumn{1}{|c}{\textbf{QWK}} & 
    \multicolumn{1}{c}{\textbf{MSE}} &
    \multicolumn{1}{|c}{\textbf{QWK}} & 
    \multicolumn{1}{c|}{\textbf{MSE}} \\
    \hline
    BDRCNNAttn [A] &
    0.302 & 0.288 &
    0.240 & 0.484 &
    0.456 & 0.476 &
    0.454 & 0.478 &
    0.426 & 0.477 &
    0.237 & 0.188 \\
    BDLSTMAttn [T] &
    0.473 & 0.240 & 
    0.305 & 0.443 & 
    0.490 & 0.455 & 
    0.564 & 0.388 & 
    0.549 & 0.389 & 
    0.402 & 0.148 \\
    MMAF &
    \textbf{0.529} & \textbf{0.229} &
    \textbf{0.334} & \textbf{0.425}&
    \textbf{0.547} & \textbf{0.406}&
    \textbf{0.586} & \textbf{0.371}&
    \textbf{0.578} & \textbf{0.371}&
    \textbf{0.435} & \textbf{0.146} \\
    \hline
    BDRCNNAttn [A]$^\dagger$ &
    0.412 &  0.318 &
    0.279 &  0.511 &
    0.455 &  0.518 &
    0.471 &  0.520 &
    0.452 &  0.499 &
    0.236 &  0.208 \\
    BDLSTMAttn [T]$^\dagger$ &
    0.543 & \textbf{0.267} &
    0.355 & 0.487 &
    0.525 & \textbf{0.469} &
    0.563 & 0.403 &
    0.564 & 0.482 &
    0.451 & 0.179 \\
    MMAF$^\dagger$ &
    \textbf{0.550} & 0.269 &
    \textbf{0.373} & \textbf{0.465} &
    \textbf{0.539} & 0.473 &
    \textbf{0.606} & \textbf{0.372} &
    \textbf{0.597} & \textbf{0.445} &
    \textbf{0.480} & \textbf{0.160} \\
    \hline
    Human-Human &
    0.676 & 0.186 &
    0.564 & 0.328 &
    0.766 & 0.241 &
    0.785 & 0.213 &
    0.823 & 0.177 &
    0.687 & 0.099 \\
    \hline
  \end{tabular}
\end{table*}

\subsection{Multi-modal with Attention Fusion (MMAF)}

In the past few years, attention mechanism  \cite{DBLP:journals/corr/BahdanauCB14} has achieved state-of-the-art on various natural language processing tasks. It enables weighting contextual information learned during each time step, allowing the model to determine which states to pay attention to. In our study, we use attention fusion \cite{Hori_2017_ICCV} to combine features from the text and audio modality for grading spoken responses (see Figure~\ref{fig:architecture}). Given bi-directional temporal hidden states from audio model (BDRCNN), $h^{a}$, and text model (BDLSTM), $h^{t}$, we combine these states, $h^{m} = [h^{a},h^{t}]$ (where $[\cdot]$ denotes
the concatenation of the state vectors) and compute context vector, $c^{m}$ as:
\begin{equation}
 e_{t} = h_{t}^{m}w_{a}; a_{t}^{m} = \frac{\exp(e_{t})}{\sum_{i=1}^{T}\exp(e_{i})}; c^{m} = \sum_{t=1}^{T}a_{t}^{m}h_{t}^{m}
\end{equation}
Here,  $h_{t}^{m}$ is the multimodal representation of the lexical/acoustic cues at time $t$ and $w_{a}$ is the weight matrix for the attention layer. The multimodal attention importance scores for each time, $a_{t}^{m}$ is obtained by multiplying the representation $h_{t}^{m}$ with the weight matrix, $w_{a}$,  followed by normalization to construct probability distribution across these cues. Finally, we calculate the context vector of cues, $c^{m}$, as a weighted summation over all time steps using multimodal attention importance scores as weights. This context vector is passed through a final dense layer before generating score for a input spoken response.

For uni-modal baselines, we applied attention weighting to individual BDRCNN and BDLSTM models (BDRCNNAttn~[A] for audio and BDLSTMAttn~[T] for text) and trained them separately for comparison.

\section{Experiments}
\label{experiments}


\subsection{Data}

In this study, we utilize data collected by Second Language Testing Inc. (SLTI) administrating Simulated Oral Proficiency Interview (SOPI) for L2 English speakers, majorly from the Philippines.
Each examinee is presented with a form consisting of six prompts of varying difficulty levels on their computer screen, and their responses are recorded. These responses are then scored independently by two expert raters. A third expert rater resolves disagreements, if any. Both the prompts and rubrics are aligned with the guidelines of the Common European Framework of Reference (CEFR) \cite{council2001common}. To answer the questions, candidates require explanatory and argumentative abilities. Depending on their performance on these tasks, they are assessed to have English oral proficiency ranging from lowest grade, A2, to highest grade, C1 (with the maximum score obtainable being equal to the difficulty of the prompt). The examinees are mostly high school graduates of age 17 years and above. Most of the recorded audios have a size lesser than 120 seconds. Table~\ref{tab:dataset} shares the other relevant statistics of the dataset. Dual-scoring with the intervention of third rater and robust rubric ensured that the scoring process is not biased towards irrelevant details like gender and features like text length. As a point of reference, the correlation between score and text length if 0.35 across all the prompts.

\subsection{Experimental setup}
\label{setup}
For each prompt, we stratify and split the responses into train, validation and test with the ratio $70:10:20$. We train the model for every prompt and report the quadratic kappa score obtained on the test set. For training the models, first, we downsample the audio responses to $16kHz$ and generate their log-scaled mel spectrograms. The number of mel bands is $128$, the length of the FFT window is $2048$ samples, and the hop length is $512$ samples.
These spectrograms are then normalized and
zero-padded to maximum response length. 
Finally, we split the output along the temporal dimension into frames of size $128 \times 128$.

We use a Deep Speech 2 \cite{DBLP:conf/icml/AmodeiABCCCCCCD16} based ASR system with trigram language model for transcription of these non-native responses. The ASR system is trained on approximately $1000$ hours of audio sampled from CommonVoice \cite{ardila2019common} and LibriSpeech dataset \cite{7178964} and further fine-tuned on approximately $22$ hours of transcribed non-native spoken responses from our dataset. This ASR system achieves a word error rate of $16.63\%$ on approximately $5$ hours of non-native spoken responses.

We tokenize the transcripts using the spacy tokenizer \cite{spacy2} and lowercase them. Words not part of training data vocabulary are mapped to the unknown token and initialized with zero vector in the embedding layer. We treat the scoring of the responses as a regression problem. The CEFR scores ($N$ levels) associated with the responses are mapped to the range $[0, N-1]$ based on the increase in proficiency level. These scores are then normalized to the range of $[0, 1]$ for training. While testing, we rescale the model output to the original score range and measure the performance.

\subsection{Evaluation metric}

For evaluating the models, we use the Quadratic Weighted Kappa (QWK) score as our primary metric. The QWK measures the agreement between two graders. It normally ranges from $0$ to $1$ and can also be negative if there is lesser agreement than expected by chance. To calculate the QWK score, a weight matrix $W$ of size $N \times N$ is constructed, where $N$ represents the number of classes, using the formula as mentioned in Eq.~\ref{Eq:2}.
\begin{equation}
\label{Eq:2}
  W_{i,j} = \frac{(i-j)^{2}}{(N-1)^{2}}
\end{equation}
Next, we construct a confusion matrix $O$ of size $N \times N$, where $O_{ij}$ is equal to the number of speech responses that receive a grade $i$ by the human and a grade $j$ by the model. 

Then we create the histogram matrix of expected grades $E$ by computing the outer product between the histogram vector of actual grades and the histogram vector of predicted grades, followed by a normalization which ensures that $E$ and $O$ have the same sum. Lastly, the QWK is obtained as in Eq.~\ref{Eq:3}.
\begin{equation}
\label{Eq:3}
  \kappa = 1 - \frac{\sum_{i,j}W_{i,j}O_{i,j}}{\sum_{i,j}W_{i,j}E_{i,j}}
\end{equation}
\subsection{Training}

We use the Adam optimization algorithm to minimize the mean squared error (MSE) loss over the training data (see Eq.~\ref{Eq:4}). 
\begin{equation}
\label{Eq:4}
  MSE(y_{true}, y_{pred}) = \frac{1}{N}\sum_{i=1}^{N}(y_{true}^{i} - y_{pred}^{i})^{2}
\end{equation}

where $N$ is the number of audio responses for a given prompt, and $y_{true}^{i}$ and $y_{pred}^{i}$ are the normalized human grade and model predicted grade for response $i$, respectively. To prevent overfitting of the model, we use dropout regularization. We train the model for a fixed number of epochs with early stopping on validation loss and select the model with the best QWK score on the validation set. 
We also optimize thresholds between each score for rounding off the predictions in order to maximize the QWK. This is done by defining the search space of thresholds between each possible score and optimizing for maximum QWK using the hyperopt package \cite{10.5555/3042817.3042832}.

\begin{table*}[th]
  \centering
  \small
  \begin{minipage}{.2\linewidth}
  \centering
  \caption{Attention split across prompts. P: Prompt Number, TA: Text Attention and AA: Audio Attention}
  \label{tab:attentionsplit}
  \begin{tabular}{|*{3}{c|}}
    \hline
    \multicolumn{1}{|c|}{\textbf{P}} & 
    \multicolumn{1}{|c|}{\textbf{TA}} & 
    \multicolumn{1}{|c|}{\textbf{AA}} \\
    \hline
	1 & 84.89\% & 15.11\% \\
    2 & 86.21\% & 13.79\% \\
    3 & 84.38\% & 15.62\% \\
    4 & 88.52\% & 11.48\% \\
    5 & 84.73\% & 15.27\% \\
    6 & 89.53\% & 10.47\% \\
    \hline
  \end{tabular}
  \end{minipage}
  \hfill
  \begin{minipage}{.4\linewidth}
  \centering
  \caption{Attention split across grades. P: Prompt Number, G: Grade, TA: Text Attention and AA: Audio Attention}
  \label{tab:attentionsplitprompt1and4}
  \begin{tabular}{|*{4}{c|}}
    \hline
    \multicolumn{1}{|c|}{\textbf{P}} & 
    \multicolumn{1}{|c|}{\textbf{G}} &
    \multicolumn{1}{|c|}{\textbf{TA}} &
    \multicolumn{1}{|c|}{\textbf{AA}} \\
    \hline
	1  & A2 & 81.63\% & 18.37\% \\
	   & Low B1 & 83.52\% & 16.48\% \\
       & High B1 & 85.44\% & 14.56\% \\
    \hline
    4  & A2 & 87.20\% & 12.80\% \\
       & Low B1 & 87.52\% & 12.48\% \\
       & High B1 & 88.26\% & 11.74\% \\
       & Low B2 & 88.97\% & 11.03\% \\
       & High B2 & 89.69\% & 10.31\% \\
    \hline
  \end{tabular}
  \end{minipage}
  \hfill
  \begin{minipage}{.3\linewidth}
  \centering
  \caption{Quadratic Kappa Score with audios replaced by white noise (denoted by $^\ast$) and by Text-to-Speech outputs of transcripts (denoted by $^\dagger$). P: Prompt Number, TO: Threshold Optimization}
  \label{tab:whitenoiseprompt1and4}
  \begin{tabular}{|*{3}{c|}}
    \hline
    \multicolumn{1}{|c|}{\textbf{P}} &
    \multicolumn{1}{|c|}{\textbf{Without TO}} &
    \multicolumn{1}{|c|}{\textbf{With TO}} \\
    \hline
	1$^\ast$  & 0.446 & 0.487 \\
    4$^\ast$  & 0.399 & 0.409 \\
    \hline
    1$^\dagger$  & 0.420 & 0.451 \\
    4$^\dagger$  & 0.388 & 0.388 \\
    \hline
  \end{tabular}
  \end{minipage}
  
\end{table*}

\section{Results and Discussion}
\label{results}

\subsection{Quantitative Evaluation}

In Table~\ref{tab:qwk}, we compare MMAF model with two strong baselines: BDRCNNAttn [A] and BDLSTMAttn [T] models on the SOPI dataset. It shows that fusing both the modalities using attention gives the best scores across all the prompts. Overall, the improvement in average QWK score across prompts for MMAF when compared to BDLSTMAttn [T] and BDRCNNAttn [A] models is close to 8.2\% and 42.6\% respectively. Optimizing thresholds improves the scores by approximately 4.8\% over text-only models and 36.5\% over audio-only model. Since prompts (1, 2 and 6) have high class imbalance and QWK is not a reliable metric since it becomes too sensitive towards the under-represented classes \cite{feinstein1990high,cicchetti1990high}, we also track MSE for comparison. We observe the average MSE across prompts for MMAF model decreases by 5.5\% and 18.3\% when compared to BDLSTMAttn [T] and BDRCNNAttn [A] models respectively.
Table~\ref{tab:qwk} shows that the performance of all models for lower difficulty level prompts (prompt 1, 2 and 6) is poor compared to higher difficulty prompts (prompt 3, 4 and 5). The human-human agreement also shows similar performance.  Table~\ref{tab:attentionsplit} shows the variation of attention scores across all the prompts. It is evident that attention weights prefer text over audio broadly in the ratio of 85:15. Due to the non-availability of any public dataset or model, we were not able to show the same results on those \cite{chen2018end,8683717}.

\begin{figure}[h]
  \centering
  \includegraphics[width=0.8\linewidth]{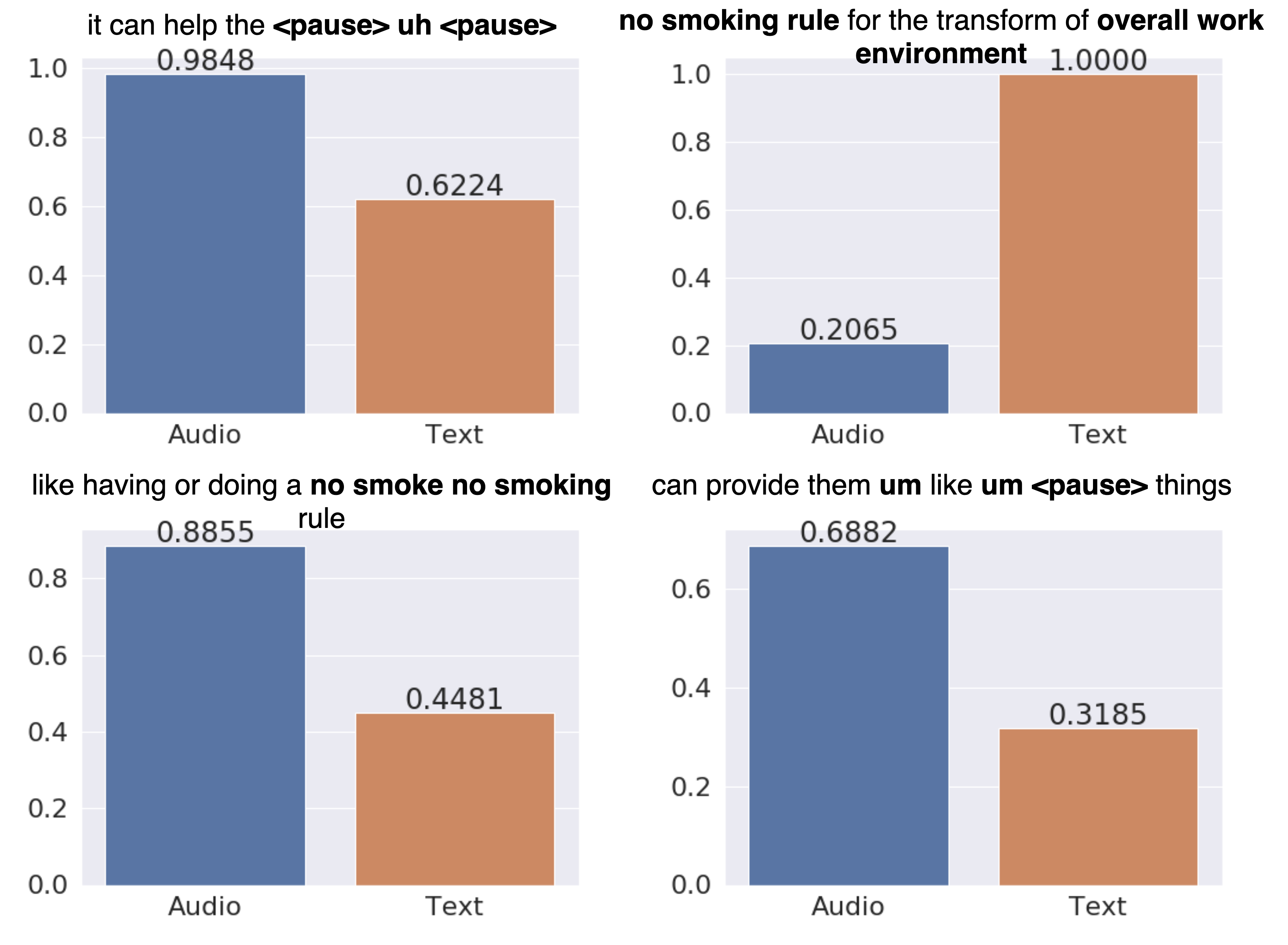}
  \caption{Attention plots of frames taken from 2 samples of Prompt~4.}
  \label{fig:pause}
\end{figure}

For further analysis, we select two exemplar prompts: one from the lower difficulty range (prompt 1) and one from the higher difficulty (prompt 4).  Table~\ref{tab:attentionsplitprompt1and4} shows that audio attention weights partially increase for low scored responses. By checking the samples, we could infer that one of the reasons for this is due to the lower intelligibility of audios for those items. This, in turn, caused ASR word error rate to go up nudging the model to partially shift the attention to audio over the text modality. \\
We also investigate the effect of audio variation on our model by performing two sets of experiments. 

\textbf{1. White noise with intact text}: Replacing audio inputs to the model with white noise, we observe that the QWK drops across both prompt 1 and 4, by an average of 23.8\% and 22\% without and with threshold optimization respectively (Table~\ref{tab:whitenoiseprompt1and4}). 

\textbf{2. Template native voice with intact text}: Next, we replace the audio inputs to the model with the Text-to-Speech outputs of the transcriptions. We use the eSpeak \cite{espeak} speech synthesizer for this experiment (see Table~\ref{tab:whitenoiseprompt1and4} for results). We observe an average drop in QWK of 27.2\% for the model without threshold optimization and of 27\% for the model with threshold optimization, across both the prompts. This performance drop can possibly be explained due to the fact that the model was originally trained on speech responses from L2 speakers, and hence performed poorly when tested on L1 speech.

Curiously, the model scored the speech with the native accented templatized audio consistently lower than the speech with white noise audio part. In our analysis, this was majorly due to two reasons: text-to-speech system did not produce much sound variation and that the generated voice was closer to a native speaker but was tested on a model trained for L2 speakers. 

\subsection{Qualitative Evaluation}

For checking the attention weights of the model on a more granular level, we manually analyzed some samples by plotting their attention weights. Figure~\ref{fig:pause} shows attention splits for one ``Low B1" and one ``Low B2" scored sample, both taken from prompt 4, which is a prompt based on the topic of smoking. We show two frames (or segmentations) for both the samples, one in each row. Each cell represents one frame of audio input and its corresponding transcription. The bar graph shows the respective audio and text attentions, normalized separately using min-max scaling across the entire response. We can observe that for filler words like \textit{``uh"} and \textit{``um"}, false starts and pauses, the model pays great attention to the audio. For content-rich fluent portions, the model chooses to pay significantly high attention to the transcripts. This is particularly visible in Figure~\ref{fig:pause} where we can see that model pays high attention to text for frames containing words like \textit{``no smoking"}, which would be important to judge the response to the question.



\section{Conclusion}
\label{conclusion}
In this paper, we investigated end-to-end deep-learning based models for the automated speech scoring task for L2 English speakers. We showed that multi-modal attention fusion works better than uni-modal networks. Results showed that transcript response of the speech is much more important as compared to audio and that it becomes progressively more important for higher scored samples. Future work would involve a feature-based analysis of all the responses and cross-question comparison of the responses on various difficulty levels. It would also be interesting to see how the ablation of network inputs and outputs fare with respect to features.

\bibliographystyle{IEEEbib}
\bibliography{strings,refs}

\begin{thebibliography}{10}

\bibitem{kumar2019get}
Yaman Kumar, Swati Aggarwal, Debanjan Mahata, Rajiv~Ratn Shah, Ponnurangam
  Kumaraguru, and Roger Zimmermann,
\newblock ``Get it scored using autosas—an automated system for scoring short
  answers,''
\newblock in {\em Proceedings of the AAAI Conference on Artificial
  Intelligence}, 2019, vol.~33, pp. 9662--9669.

\bibitem{page1966imminence}
Ellis~B Page,
\newblock ``The imminence of... grading essays by computer,''
\newblock {\em The Phi Delta Kappan}, vol. 47, no. 5, pp. 238--243, 1966.

\bibitem{page1994computer}
Ellis~Batten Page,
\newblock ``Computer grading of student prose, using modern concepts and
  software,''
\newblock {\em The Journal of experimental education}, vol. 62, no. 2, pp.
  127--142, 1994.

\bibitem{witt2000phone}
Silke~M Witt and Steve~J Young,
\newblock ``Phone-level pronunciation scoring and assessment for interactive
  language learning,''
\newblock {\em Speech communication}, vol. 30, no. 2-3, pp. 95--108, 2000.

\bibitem{doi:10.1002/ets2.12198}
Lei Chen, Klaus Zechner, Su-Youn Yoon, Keelan Evanini, Xinhao Wang, Anastassia
  Loukina, Jidong Tao, Lawrence Davis, Chong~Min Lee, Min Ma, Robert
  Mundkowsky, Chi Lu, Chee~Wee Leong, and Binod Gyawali,
\newblock ``Automated scoring of nonnative speech using the speechrater sm v.
  5.0 engine,''
\newblock {\em ETS Research Report Series}, vol. 2018, no. 1, pp. 1--31, 2018.

\bibitem{taghipour2016neural}
Kaveh Taghipour and Hwee~Tou Ng,
\newblock ``A neural approach to automated essay scoring,''
\newblock in {\em Proceedings of the 2016 conference on empirical methods in
  natural language processing}, 2016, pp. 1882--1891.

\bibitem{chen2018end}
Lei Chen, Jidong Tao, Shabnam Ghaffarzadegan, and Yao Qian,
\newblock ``End-to-end neural network based automated speech scoring,''
\newblock in {\em 2018 IEEE International Conference on Acoustics, Speech and
  Signal Processing (ICASSP)}. IEEE, 2018, pp. 6234--6238.

\bibitem{alikaniotis2016automatic}
Dimitrios Alikaniotis, Helen Yannakoudakis, and Marek Rei,
\newblock ``Automatic text scoring using neural networks,''
\newblock in {\em Proceedings of the 54th Annual Meeting of the Association for
  Computational Linguistics (Volume 1: Long Papers)}, Berlin, Germany, Aug.
  2016, pp. 715--725, Association for Computational Linguistics.

\bibitem{8683717}
Y.~{Qian}, P.~{Lange}, K.~{Evanini}, R.~{Pugh}, R.~{Ubale}, M.~{Mulholland},
  and X.~{Wang},
\newblock ``Neural approaches to automated speech scoring of monologue and
  dialogue responses,''
\newblock in {\em ICASSP 2019 - 2019 IEEE International Conference on
  Acoustics, Speech and Signal Processing (ICASSP)}, 2019, pp. 8112--8116.

\bibitem{Hori_2017_ICCV}
Chiori Hori, Takaaki Hori, Teng-Yok Lee, Ziming Zhang, Bret Harsham, John~R.
  Hershey, Tim~K. Marks, and Kazuhiko Sumi,
\newblock ``Attention-based multimodal fusion for video description,''
\newblock in {\em The IEEE International Conference on Computer Vision (ICCV)},
  Oct 2017.

\bibitem{7952585}
K.~{Choi}, G.~{Fazekas}, M.~{Sandler}, and K.~{Cho},
\newblock ``Convolutional recurrent neural networks for music classification,''
\newblock in {\em 2017 IEEE International Conference on Acoustics, Speech and
  Signal Processing (ICASSP)}, 2017, pp. 2392--2396.

\bibitem{AAAI1816431}
Yi~Tay, Minh Phan, Luu~Anh Tuan, and Siu~Cheung Hui,
\newblock ``Skipflow: Incorporating neural coherence features for end-to-end
  automatic text scoring,''
\newblock in {\em AAAI Conference on Artificial Intelligence}, 2018.

\bibitem{pennington2014glove}
Jeffrey Pennington, Richard Socher, and Christopher~D. Manning,
\newblock ``Glove: Global vectors for word representation,''
\newblock in {\em Empirical Methods in Natural Language Processing (EMNLP)},
  2014, pp. 1532--1543.

\bibitem{DBLP:journals/corr/BahdanauCB14}
Dzmitry Bahdanau, Kyunghyun Cho, and Yoshua Bengio,
\newblock ``Neural machine translation by jointly learning to align and
  translate,''
\newblock in {\em 3rd International Conference on Learning Representations,
  {ICLR} 2015, San Diego, CA, USA, May 7-9, 2015, Conference Track
  Proceedings}, 2015.

\bibitem{council2001common}
Council of~Europe. Council for Cultural Co-operation. Education Committee.
  Modern Languages~Division,
\newblock {\em Common European Framework of Reference for Languages: learning,
  teaching, assessment},
\newblock Cambridge University Press, 2001.

\bibitem{DBLP:conf/icml/AmodeiABCCCCCCD16}
Dario Amodei, Sundaram Ananthanarayanan, Rishita Anubhai, Jingliang Bai, Eric
  Battenberg, Carl Case, Jared Casper, Bryan Catanzaro, Jingdong Chen, Mike
  Chrzanowski, Adam Coates, Greg Diamos, Erich Elsen, Jesse Engel, Linxi Fan,
  Christopher Fougner, Awni~Y. Hannun, Billy Jun, Tony Han, Patrick LeGresley,
  Xiangang Li, Libby Lin, Sharan Narang, Andrew~Y. Ng, Sherjil Ozair, Ryan
  Prenger, Sheng Qian, Jonathan Raiman, Sanjeev Satheesh, David Seetapun,
  Shubho Sengupta, Chong Wang, Yi~Wang, Zhiqian Wang, Bo~Xiao, Yan Xie, Dani
  Yogatama, Jun Zhan, and Zhenyao Zhu,
\newblock ``Deep speech 2 : End-to-end speech recognition in english and
  mandarin,''
\newblock in {\em ICML}, 2016, pp. 173--182.

\bibitem{ardila2019common}
Rosana Ardila, Megan Branson, Kelly Davis, Michael Henretty, Michael Kohler,
  Josh Meyer, Reuben Morais, Lindsay Saunders, Francis~M. Tyers, and Gregor
  Weber,
\newblock ``Common voice: A massively-multilingual speech corpus,'' 2019.

\bibitem{7178964}
V.~{Panayotov}, G.~{Chen}, D.~{Povey}, and S.~{Khudanpur},
\newblock ``Librispeech: An asr corpus based on public domain audio books,''
\newblock in {\em 2015 IEEE International Conference on Acoustics, Speech and
  Signal Processing (ICASSP)}, 2015, pp. 5206--5210.

\bibitem{spacy2}
Matthew Honnibal and Ines Montani,
\newblock ``{spaCy 2}: Natural language understanding with {B}loom embeddings,
  convolutional neural networks and incremental parsing,''
\newblock To appear, 2017.

\bibitem{10.5555/3042817.3042832}
J.~Bergstra, D.~Yamins, and D.~D. Cox,
\newblock ``Making a science of model search: Hyperparameter optimization in
  hundreds of dimensions for vision architectures,''
\newblock in {\em Proceedings of the 30th International Conference on
  International Conference on Machine Learning - Volume 28}. 2013, ICML’13,
  p. I–115–I–123, JMLR.org.

\bibitem{feinstein1990high}
Alvan~R Feinstein and Domenic~V Cicchetti,
\newblock ``High agreement but low kappa: I. the problems of two paradoxes,''
\newblock {\em Journal of clinical epidemiology}, vol. 43, no. 6, pp. 543--549,
  1990.

\bibitem{cicchetti1990high}
Domenic~V Cicchetti and Alvan~R Feinstein,
\newblock ``High agreement but low kappa: Ii. resolving the paradoxes,''
\newblock {\em Journal of clinical epidemiology}, vol. 43, no. 6, pp. 551--558,
  1990.

\bibitem{espeak}
``{eSpeak}: Speech synthesizer,'' Mar 2014.

\end{thebibliography}

\end{document}